\documentclass[10pt,twocolumn,letterpaper]{article}

\pdfoutput=1

\usepackage{iccv}
\usepackage{times}
\usepackage{epsfig}
\usepackage{graphicx}
\usepackage{amsmath}
\usepackage{amssymb}
\usepackage{color}
\usepackage{transparent}
\graphicspath{{images/}}
\usepackage[font={small}]{caption,subcaption}

\iccvfinalcopy 

\usepackage{color}
\definecolor{darkred}{rgb}{0.5, 0.0, 0.0}
\definecolor{darkblue}{rgb}{0.0, 0.0, 1.0}
\newcommand{\hide}[1]{}

\newcommand{\fig}[1]{Fig.~\ref{fig:#1}}
\newcommand{\tabl}[1]{Tab.~\ref{table:#1}}
\newcommand{\sect}[1]{Sec.~\ref{sec:#1}}

\def\iccvPaperID{3780} 

\ificcvfinal\fi

\begin{document}

\title{Online Object Representations with Contrastive Learning} 

\author{S\"oren Pirk\textsuperscript{1}, Mohi Khansari\textsuperscript{2}, Yunfei Bai\textsuperscript{2}, Corey Lynch\textsuperscript{1}, Pierre Sermanet\textsuperscript{1}\\
\textsuperscript{1}Google Brain,
\textsuperscript{2}X\\
}

\maketitle

\begin{abstract}
We propose a self-supervised approach for learning representations of objects from monocular videos and
demonstrate it is particularly useful in situated settings such as robotics.
The main contributions of this paper are:
1) a self-supervising objective trained with contrastive learning that can discover and disentangle object attributes from video without using any labels;
2) we leverage object self-supervision for online adaptation: the longer our online model looks at objects in a video, the lower the object identification error, while the offline baseline remains with a large fixed error;
3) to explore the possibilities of a system entirely free of human supervision, we let a robot collect its own data, train on this data with our self-supervise scheme, and then show the robot can point to objects similar to the one presented in front of it, demonstrating generalization of object attributes.
An interesting and perhaps surprising finding of this approach is that given a limited set of objects, object correspondences will naturally emerge when using contrastive learning without requiring explicit positive pairs.
Videos illustrating online object adaptation and robotic pointing are available at: https://online-objects.github.io/.\vspace{2mm}
\end{abstract}

\begin{figure}[ht]
  \begin{center}
  \includegraphics[width=1.0\linewidth]{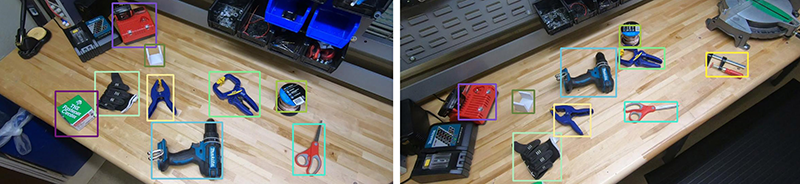}\\[2mm]
  \includegraphics[width=1.0\linewidth]{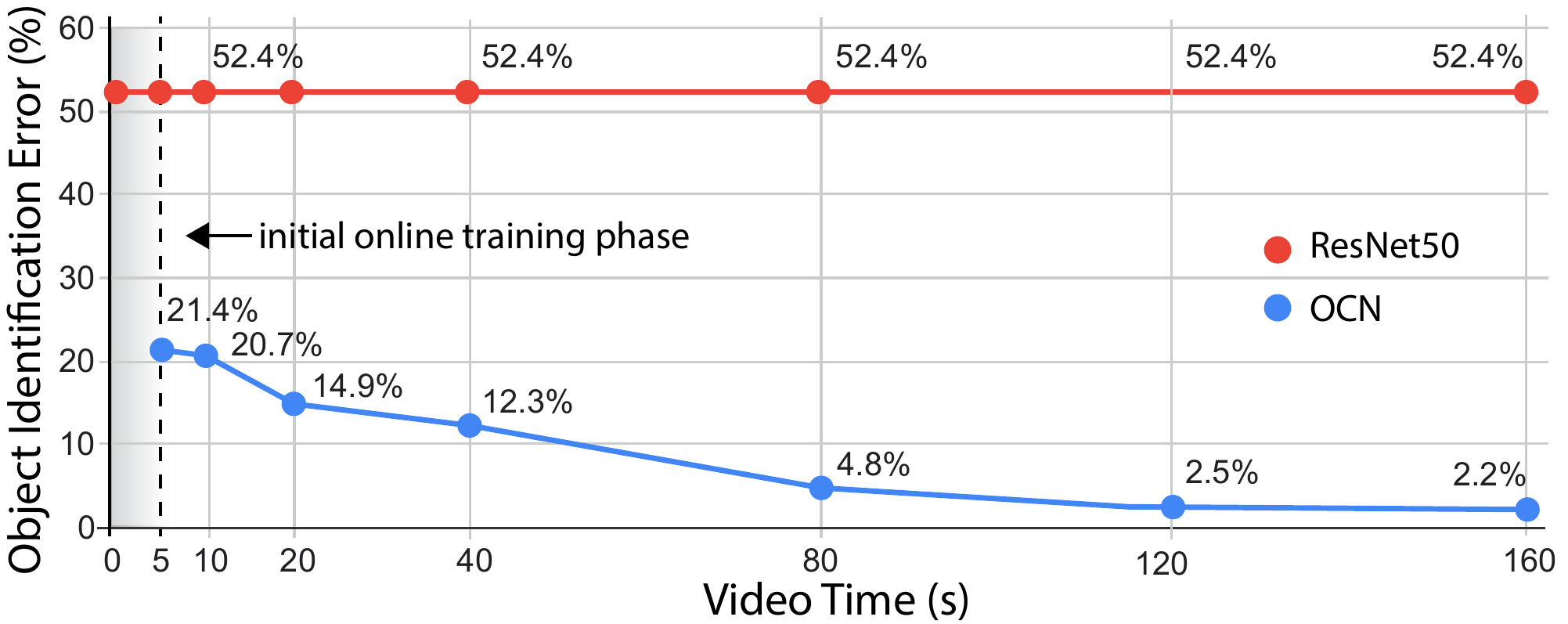}
  \end{center}
    \vspace{-6mm}
  \caption{\textbf{The longer our model looks at objects in a video, the lower the object identification error.} Top: example frames of a work bench video along with the detected objects. Bottom:~result of online training on the same video. Our model self-supervises object representations as the video progresses and converges to 2\% error while the offline baseline remains at 52\% error.}
  \label{fig:online}
  \vspace{-5mm}
\end{figure}

\vspace{-5mm}
\section{Introduction}
One of the biggest challenges in real world robotics is robustness and adaptability to new situations. A robot deployed in the real world is likely to encounter a number of objects it has never seen before. Even if it can identify the class of an object, it may be useful to recognize a particular instance of it.
Relying on human supervision in this context is unrealistic.
Instead if a robot can self-supervise its understanding of objects, it can adapt to new situations when using online learning.
Online self-supervision is key to robustness and adaptability and arguably a prerequisite to real-world deployment. Moreover, removing human supervision has the potential to enable learning richer and less biased continuous representations than those obtained by supervised training and a limited set of discrete labels. Unbiased representations can prove useful in unknown future environments different from the ones seen during supervision, a typical challenge for robotics. Furthermore, the ability to autonomously train to recognize and differentiate previously unseen objects as well as to infer general properties and attributes is an important skill for robotic agents.

In this work we focus on situated settings (i.e. an agent is embedded in an environment), which allows us
to use temporal continuity as the basis for self-supervising correspondences between different views of objects.
We present a self-supervised method that learns representations to disentangle perceptual and semantic object attributes such as class, function, and color. We automatically acquire training data by capturing videos with a real robot; a robot base moves around a table to capture objects in various arrangements. Assuming a pre-existing objectness detector, we extract objects from random frames of a scene containing the same objects, and let a metric learning system decide how to assign positive and negative pairs of embeddings. Representations that generalize across objects naturally emerge despite not being given groundtruth matches. Unlike previous methods, we abstain from employing additional self-supervisory training signals such as depth or those used for tracking. The only input to the system are monocular videos. This simplifies data collection and allows our embedding to integrate into existing end-to-end learning pipelines. We demonstrate that a trained Object-Contrastive Network~(OCN) embedding allows us to reliably identify object instances based on their visual features such as color and shape. Moreover, we show that objects are also organized along their semantic or functional properties. For example, a cup might not only be associated with other cups, but also with other containers like bowls or vases.

Fig.~\ref{fig:online} shows the effectiveness of online self-supervision: by training on randomly selected frames of a continuous video sequence (top) OCN can adapt to the present objects and thereby lower the object identification error. While the supervised baseline remains at a constant high error rate~(52.4\%), OCN converges to a 2.2\% error. The graph (bottom) shows the object identification error obtained by training on  progressively longer sub-sequences of a 200 seconds video.

The key contributions of this work are:
(1)~a self-supervising objective trained with contrastive learning that can discover and disentangle object attributes from video without using any labels;
(2) we leverage object self-supervision for online adaptation: the longer our online model looks at objects in a video, the lower the object identification error, while the offline baseline remains with a large fixed error;
(3)~to explore the possibilities of a system entirely free of human supervision: we let a robot collect its own data, then train on this data with our self-supervised training scheme, and show the robot can point to objects similar to the one presented in front of it, demonstrating generalization of object attributes.

\vspace{2mm}
\section{Related Work}

\textbf{Object discovery from visual media.} Identifying objects and their attributes has a long history in computer vision and robotics~\cite{Tuytelaars09}. Traditionally, approaches focus on identifying regions in unlabeled images to locate and identify objects~\cite{1541280,DBLP:conf/cvpr/RussellFESZ06,4270036, 4587803,4587502}. Discovering objects based on the notion of 'objectness' instead of specific categories enables more principled strategies for object recognition~\cite{UijlingsIJCV2013, Romea-2011-7355}. Several methods address the challenge to discover, track, and segment objects in videos based on supervised~\cite{42961} or unsupervised~\cite{kwak2015,18cb3eb6fa104c8da9fcaccb96837a5b,Haller_2017_ICCV} techniques. The spatio-temporal signal present in videos can also help to reveal additional cues that allow to identify objects~\cite{Wang_UnsupICCV2015,DBLP:conf/cvpr/JainXG17}. In the context of robotics, methods also focus on exploiting depth to discover objects and their properties~\cite{6225107,Karpathy_ICRA2013}.

Many recent approaches exploit the effectiveness of convolutional deep neural networks to detect objects~\cite{NIPS2015_5638,44872,Lin2017} and to even provide pixel-precise segmentations~\cite{he2017maskrcnn}. While the detection efficiency of these methods is unparalleled, they rely on supervised training procedures and therefore require large amounts of labeled data. Self-supervised methods for the discovery of object attributes mostly focus on learning representations by identifying features in multi-view imagery~\cite{DBLP:journals/corr/abs-1712-07629,7299135} and videos~\cite{Wang_UnsupICCV2015}, or by stabilizing the training signal through domain randomization~\cite{doersch2015unsupervised,zhang2018mixup}. 

Some methods not only operate on RGB images but also employ additional signals, such as depth~\cite{florence2018,Pot2018SelfsupervisorySF} or egomotion~\cite{LSM2015} to self-supervise the learning process. It has been recognized, that contrasting observations from multiple views can provide a view-invariant training signal allowing to even differentiate subtle cues as relevant features that can be leveraged for instance categorization and imitation learning tasks~\cite{Sermanet2017TCN}.  

\textbf{Unsupervised representation learning.} Unlike supervised learning techniques, unsupervised methods focus on learning representations directly from data to enable image retrieval~\cite{7410376}, transfer learning~\cite{zhang2017split}, image denoising~\cite{Vincent:2008:ECR:1390156.1390294}, and other tasks~\cite{journals/corr/DumoulinBPLAMC16,DBLP:journals/corr/KumarCR15}.
Using data from multiple modalities, such as imagery of multiple views~\cite{Sermanet2017TCN}, sound~\cite{owens16a,aytar2016soundnet}, or other sensory inputs~\cite{s17122735}, along with the often inherent spatio-temporal coherence~\cite{doersch2015unsupervised,DBLP:journals/corr/RadfordMC15}, can facilitate the unsupervised learning of representations and embeddings. For example, ~\cite{Zagoruyko_2015_CVPR} explore multiple architectures to compare image patches and ~\cite{pathakCVPR17learning} exploit temporal coherence to learn object-centric features. ~\cite{DBLP:journals/corr/GaoJG16} rely of spatial proximity of detected objects to determine attraction in metric learning, OCN operates similarly but does not require spatial proximity for positive matches, it does however take advantage of the likely presence of a same object in any pair of frames within a video. \cite{DBLP:journals/corr/abs-1710-02139} also take a similar unsupervised metric learning approach for tracking specific faces, using tracking trajectories and heuristics for matching trajectories and obtain richer positive matches. While our approach is simpler in that it does not require tracking or 3D matching, it could be augmented with extra matching signals.

In robotics and other real-world scenarios where agents are often only able obtain sparse signals from their environment, self-learned embeddings can serve as an efficient representation to optimize learning objectives. \cite{pathakICMl17curiosity} introduce a curiosity-driven approach to obtain a reward signal from visual inputs; other methods use similar strategies to enable grasping~\cite{7487517} and manipulation tasks~\cite{Sermanet2017TCN}, or to be pose and background agnostic~\cite{Held2015DeepLF}. \cite{mitash2017self} jointly uses 3D synthetic and real data to learn a representation to detect objects and estimate their pose, even for cluttered configurations. \cite{DBLP:journals/corr/abs-1801-08985} learn semantic classes of objects in videos by integrating clustering into a convolutional neural network. 

\begin{figure*}[ht]
  \begin{center}
  \includegraphics[width=1.0\linewidth]{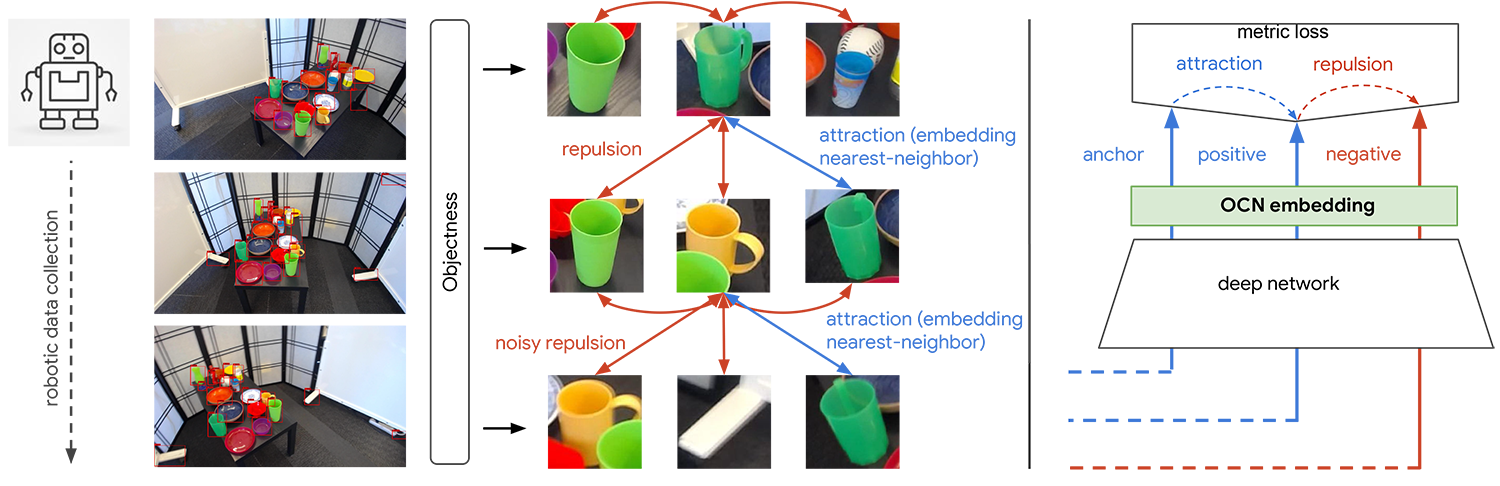} 
  \end{center}
  \vspace{-7mm}
  \caption{\textbf{Object-Contrastive Networks (OCN)}: by attracting nearest neighbors in embedding space and repulsing others using metric learning, continuous object representations naturally emerge. In a video collected by a robot looking at a table from different viewpoints, objects are extracted from random pairs of frames. Given two lists of objects, each object is attracted to its closest neighbor while being pushed against all other objects. Noisy repulsion may occur when the same object across viewpoint is not matched against itself. However the learning still converges towards disentangled and semantically meaningful object representations. }
  \label{fig:teaser}  
  \vspace{-3mm}
\end{figure*}

\section{Learning of Object Representations}

We propose a model called Object-Contrastive Network~(OCN) trained with a metric learning loss (\fig{teaser}).
The approach is very simple: 1) extract object bounding boxes using a general off-the-shelf objectness detector \cite{NIPS2015_5638}, 2) train a deep object model on each cropped image extracted from any random pair of frames from the video, using the following training objective: nearest neighbors in the embedding space are pulled together from different frames while being pushed away from the other objects from any frame (using n-pairs loss~\cite{NIPS2016_6200}). This does not rely on knowing the true correspondence between objects.
The fact that this works at all despite not using any labels might be surprising.
One of the main findings of this paper is that given a limited set of objects, object correspondences will naturally emerge when using metric learning.
One advantage of self-supervising object representation is that these continuous representations are not biased by or limited to a discrete set of labels determined by human annotators.
We show these embeddings discover and disentangle object attributes and generalize to previously unseen environments.

We propose a self-supervised approach to learn object representations for the following reasons: (1) make data collection simple and scalable, (2) increase autonomy in robotics by continuously learning about new objects without assistance, (3) discover continuous representations that are richer and more nuanced than the discrete set of attributes that humans might provide as supervision which may not match future and new environments. 
All these objectives require a method that can learn about objects and differentiate them without supervision. To bootstrap our learning signal we leverage two assumptions: (1) we are provided with a general objectness model so that we can attend to individual objects in a scene, (2) during an observation sequence the same objects will be present in most frames (this can later be relaxed by using an approximate estimation of ego-motion). Given a video sequence around a scene containing multiple objects, we randomly select two frames $I$ and $\hat{I}$ in the sequence and detect the objects present in each image. Let us assume $N$ and $M$ objects are detected in image $I$ and $\hat{I}$, respectively. Each of the $n$-th and $m$-th cropped object images are embedded in a low dimensional space, organized by a metric learning objective. Unlike traditional methods which rely on human-provided similarity labels to drive metric learning, we use a self-supervised approach to mine synthetic similarity labels.

\subsection{Objectness Detection}
\label{sec:object_detection}
To detect objects, we use Faster-RCNN~\cite{NIPS2015_5638} trained on the COCO object detection dataset~\cite{10.1007/978-3-319-10602-1_48}. Faster-RCNN detects objects in two stages: first generate class-agnostic bounding box proposals of all objects present in an image (Fig.~\ref{fig:teaser}), second associate detected objects with class labels. We use OCN to discover object attributes, and only rely on the first \textit{objectness} stage of Faster-R-CNN to detect object candidates. 

\subsection{Metric Loss for Object Disentanglement}
\label{lab:metric_loss}
We denote a cropped object image by $x \in \mathcal{X}$ and compute its embedding based on a convolutional neural network $f(x): \mathcal{X} \rightarrow K$.
Note that for simplicity we may omit $x$ from $f(x)$ while $f$ inherits all superscripts and subscripts. Let us consider two pairs of images $I$ and $\hat{I}$ that are taken at random from the same contiguous observation sequence. Let us also assume there are $n$ and $m$ objects detected in $I$ and $\hat{I}$ respectively. We denote the $n$-th and $m$-th objects in the images $I$ and $\hat{I}$ as $x_n^{I}$ and $x_m^{\hat{I}}$, respectively. We compute the distance matrix $D_{n,m} = \sqrt{(f_{n}^{I} - f_{m}^{\hat{I}}})^2,~n\in1..N,~m\in1..M$. For every embedded \textit{anchor} $f_{n}^{I},~n\in1..N$, we select a \textit{positive} embedding $f_{m}^{\hat{I}}$ with minimum distance as \textit{positive}: $f_{n+}^{\hat{I}} = argmin(D_{n,m})$. %
Given a batch of (\textit{anchor}, \textit{positive}) pairs $\{(x_i, x_i^+)\}_{i=1}^N$, the n-pair loss is defined as follows~\cite{NIPS2016_6200}: 
\vspace{-1mm}
\begin{align*} 
\mathcal{L}_{N-pair}\big(\{(x_i, x_i^+)\}_{i=1}^N;f\big) = \\
\frac{1}{N} \sum_{i=1}^N log \Big(1 + \sum_{j \neq i} exp(f_i^\intercal f_j^+ - f_i^\intercal f_i^+) \Big)
\end{align*} 
\vspace{-3mm}

The loss learns embeddings that identify ground truth (anchor, positive)-pairs from all other (anchor, negative)-pairs in the same batch. It is formulated as a sum of softmax multi-class cross-entropy losses over a batch, encouraging the inner product of each (anchor, positive)-pair ($f_i$, $f_i^+$) to be larger than all (anchor, negative)-pairs ($f_i$, $f_{j\neq i}^+$). The final OCN training objective over a sequence is the sum of npairs losses over all pairs of individual frames:
\vspace{-1mm}
\begin{align*} 
\mathcal{L}_{OCN} = \mathcal{L}_{N-pair}\big(\{(x_n^{I}, x_{n+}^{\hat{I}})\}_{n=1}^N;f\big) \\
                  + \mathcal{L}_{N-pair}\big(\{(x_m^{\hat{I}}, x_{m+}^{I})\}_{m=1}^M;f\big)
\end{align*}                   
\vspace{-1mm}
\subsection{Architecture and Embedding Space}

\begin{figure}[h] 
  \vspace{-2mm}
  \begin{center}
  \includegraphics[width=1.0\linewidth]{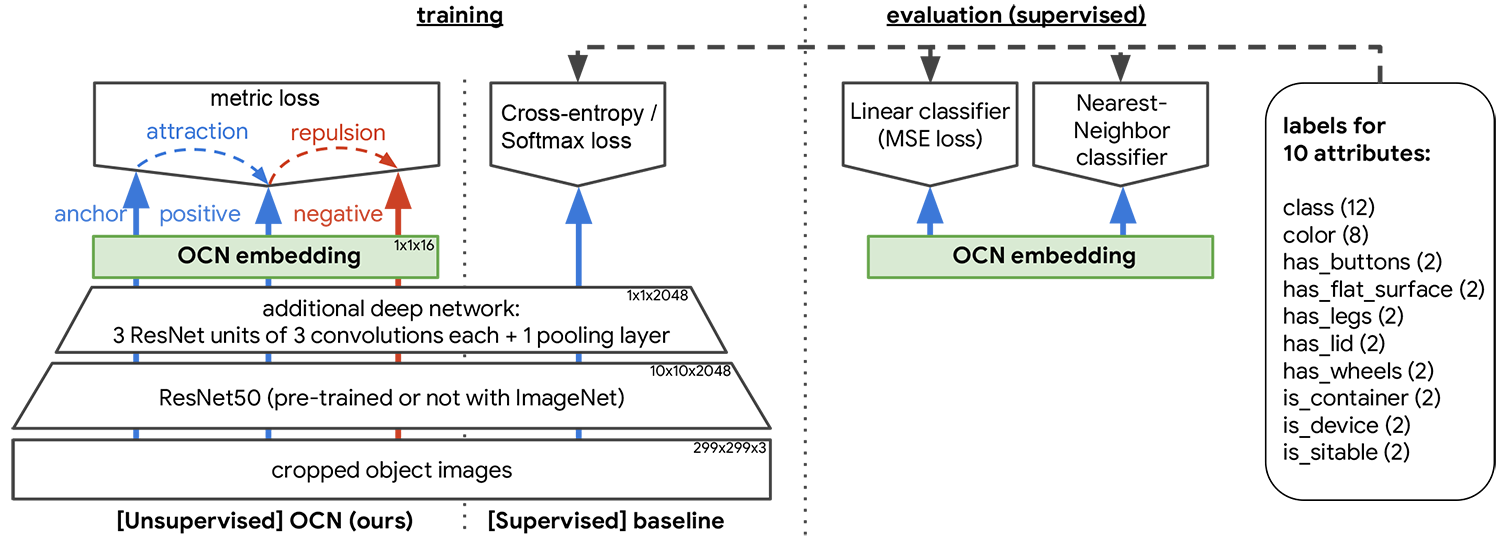} 
  \end{center}
  \vspace{-4mm}
  \caption{\textbf{Models and baselines:} for comparison purposes all models evaluated in \sect{results} share the same architecture of a standard ResNet50 model followed by additional layers. While the architectures are shared, the weights are not across models. While the unsupervised model (left) does not require supervision labels, the 'softmax' baseline as well as the supervised evaluations (right) use attributes labels provided with each object. We evaluate the quality of the embeddings with two types of classifiers: linear and nearest neighbor.}
  \label{fig:models}  
  \vspace{-2mm}
\end{figure}

OCN takes a standard ResNet50 architecture until layer \textit{global\_pool} and initializes it with ImageNet pre-trained weights. We then add three additional ResNet convolutional layers and a fully connected layer to produce the final embedding. 
The network is trained with the n-pairs metric learning loss as discussed in Sec.~\ref{lab:metric_loss}. Our architecture is depicted in \fig{teaser} and \fig{models}.

\textbf{Object-centric Embeding Space:} By using multiple views of the same scene and by attending to individual objects, our architecture allows us to differentiate subtle variations of object attributes. Observing the same object across different views facilitates learning invariance to scene-specific properties, such as scale, occlusion, lighting, and background, as each frame exhibits variations of these factors. The network solves the metric loss by representing object-centric attributes, such as shape, function, or color, as these are consistent for (anchor, positive)-pairs, and dissimilar for (anchor, negative)-pairs.  

\subsection{Discussion}

One might expect that this approach may only work if it is given an initialization so that matching the same object across multiple frames is more likely than random chance. While ImageNet pretraining certainly helps convergence as shown in \tabl{classification}, it is not a requirement to learn meaningful representations as shown in \sect{random_weights}. When all weights are random and no labels are provided, what can drive the network to consistently converge to meaningful embeddings? We estimate that the co-occurrence of the following hypotheses drives this convergence: (1)~objects often remains visually similar to themselves across multiple viewpoints, (2)~limiting the possible object matches within a scene increases the likelihood of a positive match, (3)~the low-dimensionality of the embedding space forces the model to generalize by sharing abstract features across objects, (4)~the smoothness of embeddings learned with metric learning facilitates convergence when supervision signals are weak, and (5)~occasional true-positive matches (even by chance) yield more coherent gradients than false-positive matches which produce inconsistent gradients and dissipate as noise, leading over time to an acceleration of consistent gradients and stronger initial supervision signal.

\section{Experiments}

\textbf{Online Results:}
we quantitatively evaluate the online adaptation capabilities of our model through the object identification error of entirely novel objects.
In \fig{online} we show that a model observing objects for a few minutes from different angles can self-teach to identify them almost perfectly while the offline supervised approach cannot. OCN is trained on the first 5, 10, 20, 40, 80, and 160 seconds of the 200 seconds video, then evaluated on the identification error of the last 40 seconds of the video for each phase. The supervised offline baseline stays at a 52.4\% error, while OCN improves down to 2\% error after 80s, a 25x error reduction.

\textbf{Robotic Experiments:}
here we let a robot collect its own data by looking at a table from multiple angles (\fig{teaser} and \fig{real_dataset}).
It then trains itself with OCN on that data, and is asked to point to objects similar to the one presented in front of it.
Objects can be similar in terms of shape, color or class.
If able to perform that task, the robot has learned to distinguish and recognize these attributes entirely on its own, from scratch and by collecting its own data. We find in \tabl{pointing_evaluation} that the robot is able to perform the pointing task with 72\% recognition accuracy of 5 classes, and 89\% recognition accuracy of the binary is-container attribute.

\textbf{Offline Analysis:}
to analyze what our model is able to disentangle, we quantitatively evaluate performance on a large-scale synthetic dataset with 12k object models (e.g. \fig{object_query0}), as well as on a real dataset collected by a robot and show that our unsupervised object understanding generalizes to previously unseen objects. In \tabl{classification} we find that our self-supervised model closely follows its supervised equivalent baseline when trained with metric learning. As expected the cross-entropy/softmax supervised baseline approach performs best and establishes the error lower bound while the ResNet50 baseline are upper-bound results.


\section{Data Collection and Training}

We generated three datasets of real and synthetic objects for our experiments. For the real data we arrange objects in table-top configurations and use frames from continuous camera trajectories. The labeled synthetic data is generated from renderings of 3D objects in a similar configuration. Details about the datasets are reported in Tab.~\ref{tab:data_set}.

\subsection{Real Data for Online Training}
\label{subsec:online_training_data}

\begin{figure}[t]
  \begin{center}
  \includegraphics[width=\linewidth]{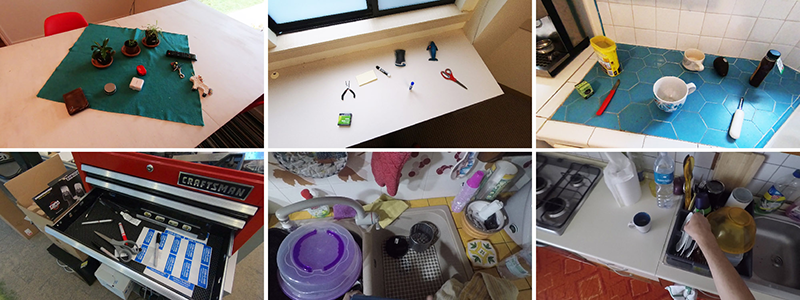}
  \end{center}
  \vspace{-6mm}
  \caption{Six of the environments we used for our self-supervised online experiment. Top: living room, office, kitchen. Bottom: one of our more challenging scenes, and two examples of the Epic-Kitchens~\cite{Damen2018EPICKITCHENS} dataset.}
  \label{fig:online_dataset}
\vspace{-1mm}
  \begin{center}
  \includegraphics[width=\linewidth]{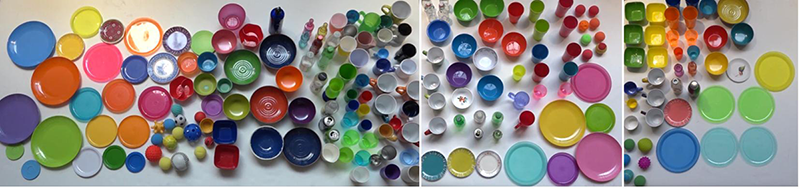}
  \end{center}
  \vspace{-6mm}
  \caption{We use 187 unique object instance in the real world experiments: 110 object for training (left), 43 objects for test (center), and 34 objects for validation (right). The degree of similarity makes it harder to differentiate these objects.}
  \label{fig:real_dataset}
  \vspace{-3mm}
\end{figure}

For the online adaptation experiment, we captured videos of table-top object configurations in the 5 environments (categories): kids room, kitchen, living room, office, and work bench (Figs.~\ref{fig:online}, ~\ref{fig:online_dataset}, and \ref{fig:online_result}). We show objects common to each environment (e.g. toys for kids room, tools for work bench) and arrange them randomly; we captured 3 videos for each environment and used 75 unique objects. To allow capturing the objects from multiple view points we use a head-mounted camera and interact with the objects (e.g. turning or flipping them). Additionally, we captured 5 videos of more challenging object configurations (referred to as `challenging') with cluttered objects or where objects are not permanently in view. Finally, we selected 5 videos from the Epic-Kitchens~\cite{Damen2018EPICKITCHENS} dataset to show that OCN can also operate on even more realistic video sequences.

\begin{figure*}[t]

  \begin{center}
  \includegraphics[width=\linewidth]{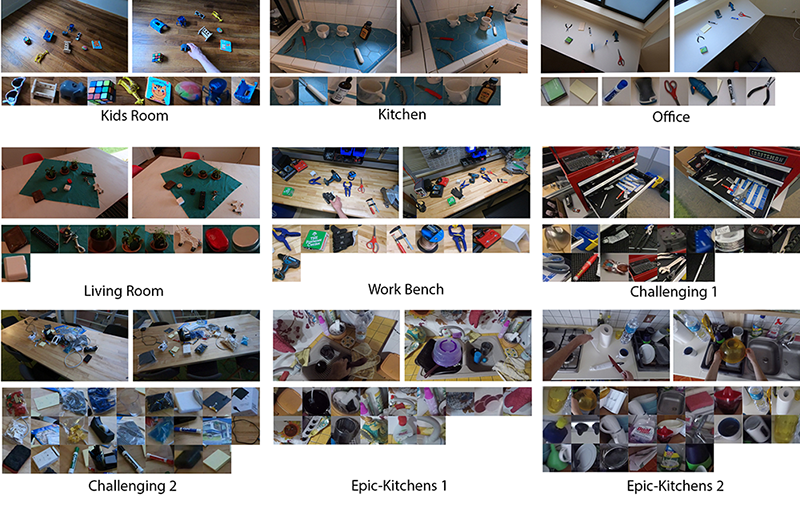}
  \end{center}
  \vspace{-8mm}
  \caption{Example frames from the sequences of the 7 categories: kids room, kitchen, office, living room, work bench, `challenging', and Epic-Kitchens. For each video we show the manually selected reference objects used for the  object identification experiment.}
  \label{fig:all_videos}
\end{figure*}

From all these videos we take the first 200 seconds and sample the sequence with 15 FPS to extract 3,000 frames. We then use the first 2,400 frames~(160s) for training OCN and the remaining 600 frames~(40s) for evaluation. We manually select up to 30 reference objects (those we  interacted with) as cropped images for each video in order of their appearance from the beginning of the video (\fig{all_videos}). Then we use object detection to find the bounding boxes of these objects in the video sequence and manually correct these boxes (add, delete) in case object detection did not identify an object. This allows us to prevent artifacts of the object detection to interfere with the evaluation of OCN..

\subsection{Automatic Real Data Collection}

\begin{figure}[ht]
    \begin{centering}
  \includegraphics[width=\linewidth]{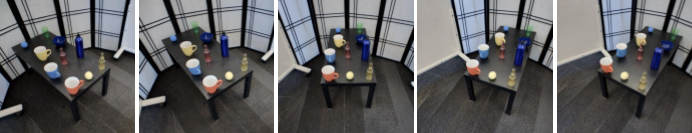}
  \end{centering}
  \vspace{-6mm}
  \caption{Consecutive frames captured with our robotic setup. At each run we randomly select 10 objects and place them on the table. Then a robot moves around the table and take snapshots of the table at different angles. We collect in average 80-120 images per scene. We select pairs of two frames of the captured trajectory and train the OCN on the detected objects.}
  \label{fig:real_data_collection}
    \vspace{3mm}
  \begin{centering}
  \includegraphics[width=\linewidth]{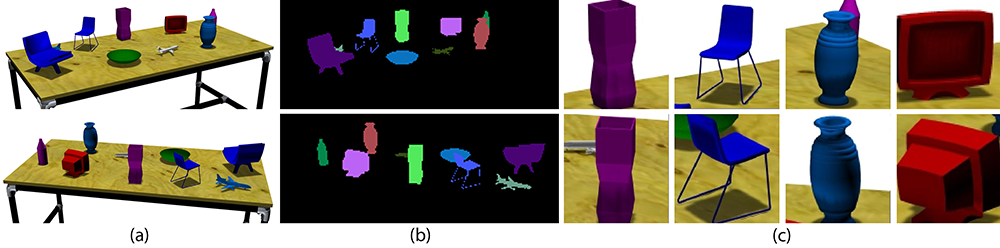}
  \end{centering}
  \vspace{-6mm}
  \caption{Synthetic data: two frames of a synthetically generated scene of table-top objects (a) and a subset of the detected objects (c). To validate our method against a supervised baseline, we additionally render color masks (b) that allow us to identify objects across the views and to associate them with their semantic attributes after object detection. Note that objects have the same color id across different views. The color id's allow us to supervise the OCN during training.}
  \label{fig:synthetic_dataset}
\end{figure}

To explore the possibilities of a system entirely free of human supervision we automated the real world data collection by using a mobile robot equipped with an HD camera (Fig.~\ref{fig:real_data_collection}). For this dataset we use 187 unique object instances spread across six categories including `balls', `bottles \& cans', `bowls', `cups \& mugs', `glasses', and `plates'. \tabl{real_data_set} provides details about the number of objects in each category and how they are split between training, test, and validation. Note that we distinguish between cups \& mugs and glasses categories based on whether it has a handle. Fig.~\ref{fig:real_dataset} shows our entire object dataset.

At each run, we place about 10 objects on the table and then trigger the capturing process by having the robot rotate around the table by 90 degrees (Fig.~\ref{fig:real_data_collection}). On average 130 images are captured at each run. We select random pairs of frames from each trajectory for training OCN. We performed 345, 109, and 122 runs of data collection for training, test, and validation dataset. In total 43,084 images were captured for OCN training and 15,061 and 16,385 were used for test and validation, respectively.

\label{sec:synthetic_data}
\subsection{Synthetic Data Generation}
\label{sec:synthetic_data_generation}

To generate diverse object configurations we use 12 categories (airplane, car, chair, cup, bottle, bowl, guitars, keyboard, lamp, monitor, radio, vase) from  ModelNet~\cite{conf/cvpr/WuSKYZTX15}. The selected categories cover around 8k models of the 12k models available in the entire dataset. ModelNet provides the object models in a 80-20 split for training and testing. We further split the testing data into models for test and validation, resulting in a 80-10-10 split for training, validation, and test. For validation purposes, we manually assign each model labels describing the semantic and functional properties of the object, including the labels `class', `has lid', `has wheels', `has buttons', `has flat surface', `has legs', `is container', `is sittable', `is device'.

We randomly define the number of objects (up to 20) in a scene  (Fig.~\ref{fig:synthetic_dataset}).
Further, we randomly define the positions of the objects and vary their sizes, both so that they do not intersect. Additionally, each object is assigned one of eight predefined colors. We use this setup to generate 100K scenes for training, and 50K scenes for each, validation and testing. For each scene we generate 10 views and select random combination of two views for detecting objects. In total we produced 400K views (200K pairs) for training and 50K views (25K pairs) for each, validation and testing.

\subsection{Training}

OCN is trained based on two views of the same synthetic or real scene. We randomly pick two frames of a video sequence and detect objects to produce two sets of cropped images.
The distance matrix $D_{n,m}$ (Sec.~\ref{lab:metric_loss}) is constructed based on the individually detected objects for each of the two frames. The object detector was not specifically trained on any of our datasets.

As the number of detected objects per view varies, we reciprocally use both frames to find anchors and their corresponding positives as discussed in~Sec.~\ref{lab:metric_loss}. Across our experiments, we observed an embeddings size of 32-64 provides optimal results; the OCN training converged after 600k-1.2M iterations.

\section{Experimental Results}
\label{sec:results}

We evaluated the effectiveness of OCN embeddings on identifying objects through self-supervised online training, a real robotics pointing tasks, and  large-scale synthetic data.

\subsection{Online Object Identification}

Our self-supervised online training scheme enables to train and to evaluate on unseen objects and scenes. This is of utmost importance for robotic agents to ensure adaptability and robustness in real world scenes. To show the potential of our method for these situations we use OCN embeddings to identify instances of objects across multiple views and over time.
\begin{figure}[t]
  \begin{center}
  \includegraphics[width=\linewidth]{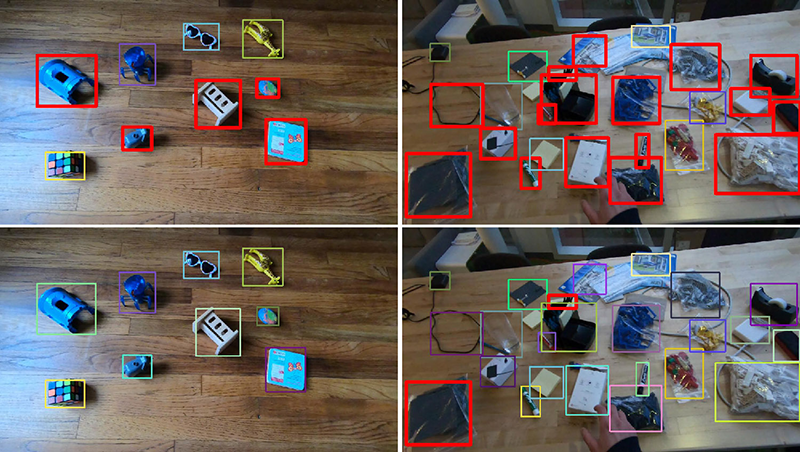}
  \end{center}
  \vspace{-5mm}
  \caption{Comparison of identifying objects with ResNet50 (top) and OCN (bottom) embeddings for the environments kids room~(left) and challenging (right). Red bounding boxes indicate a mismatch of the ground truth and associated index.}
  \label{fig:online_result}
  \vspace{-1mm}

  \begin{center}
  \includegraphics[width=\linewidth]{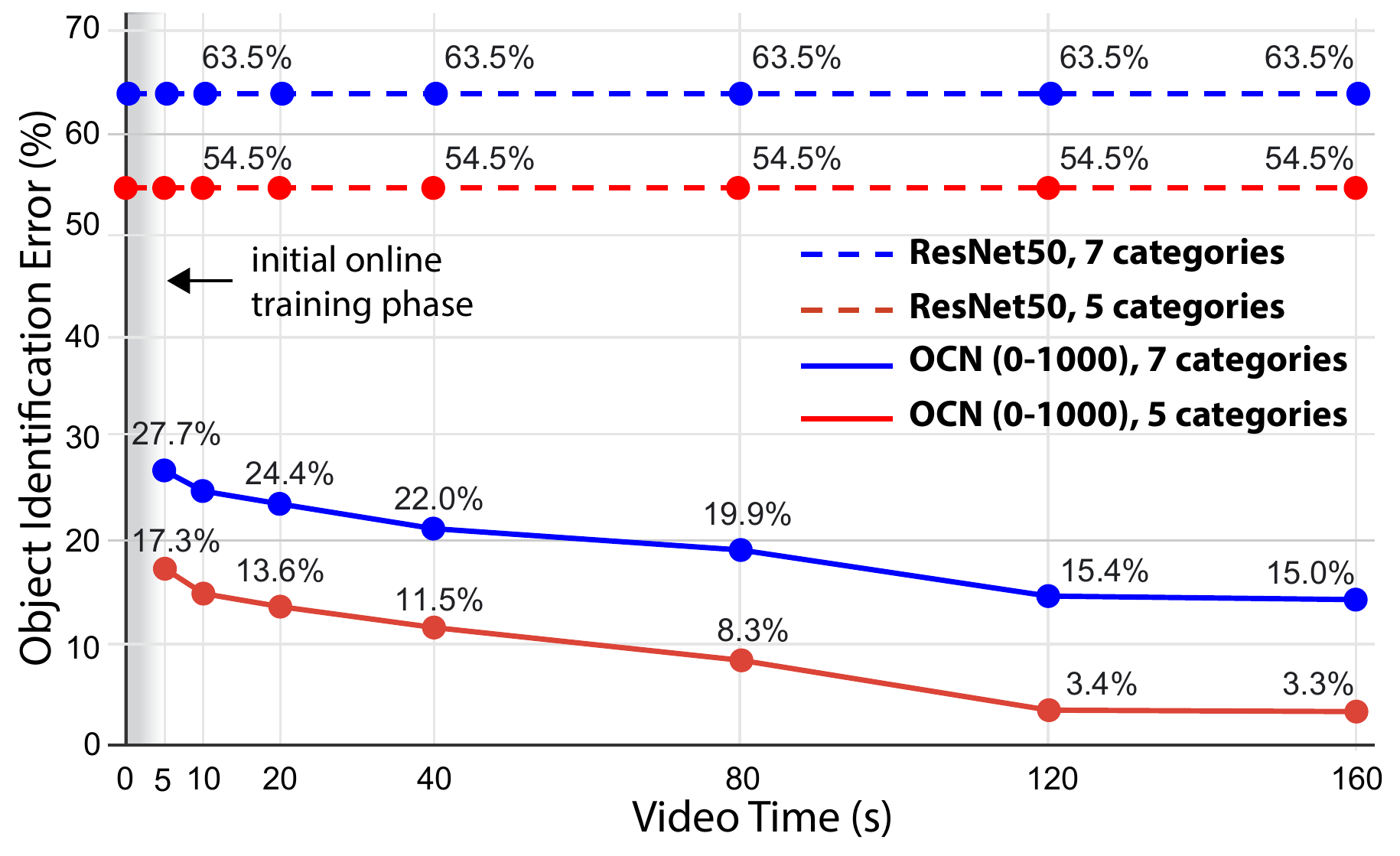}
  \end{center}
  \vspace{-6mm}
  \caption{Evaluation of online adaptation: we train an OCN on the first 5, 10, 20, 40, 80, and 160 seconds of each 200 second test video and then evaluate on the remaining 40 seconds. Here we report the lowest average error of all videos (over 1000K iterations) of online adaptation. Results are shown for 5 and 7 categories and compared to the ResNet50 baseline.}
  \label{fig:avg_all}
  \vspace{-5mm}
\end{figure}

We use sequences of videos showing objects in random configurations in different environments (Sec.~\ref{subsec:online_training_data}, Fig.~\ref{fig:online_dataset}) and train an OCN on the first 5, 10, 20, 40, 80, and 160 seconds of a 200 seconds video. Our dataset provides object bounding boxes and unique identifiers for each object as well as reference objects and their identifiers. The goal of this experiment is to assign the identifier of a reference object to the matching object detected in a video frame. We evaluate the identification error (ground truth index vs. assigned index) of objects present in the last 40 seconds of each video and for each training phase to then compare our results to a ResNet50 (2048-dimensional vectors) baseline.

We train an OCN for each video individually. Therefore, we only split our dataset into validation and testing data. For the categories kids room, kitchen, living room, office, and work bench we use 2 videos for validation and 1 video for testing; for the categories `challenging' and epic kitchen we use 3 videos for validation and 2 for testing. We jointly train on the validation videos to find meaningful hyperparameters across the categories and use the same hyperparameters for the test videos.

Fig.~\ref{fig:online_result} shows the same video frames of two scenes from our dataset. Objects with wrongly matched indices are shown with a red bounding box, correctly matched objects are shown with random colors.
In \fig{avg_all} and Tab.~\ref{tab:online_eval} we report the average error of OCN object identification across our videos compared to the ResNet50 baseline. As the supervised model cannot adapt to unknown objects OCN outperforms this baseline by a large margin. Furthermore, the optimal result among the first 50K training iterations closely follows the overall optimum obtained after 1000K iterations. We report results for 5 categories (kids room, kitchen, living room, office, work bench), that we specifically captured for evaluating OCN and the whole dataset (7 categories). The latter data also shows cluttered objects which are more challenging to detect. 
To evaluate the degree of how object detection is limiting application of OCN we counted the number of manually added bounding boxes of the evaluation sequences. On average the evaluation sequences of the 5 categories have 5,122 boxes (468 added, 9.13\%), while the whole dataset (7 categories) has 5,002 boxes on average~(1183 added, 25.94\%).

\begin{table}[h]
\begin{center}
\caption{Evaluation of online adaptation: we report the lowest error among 50K and 1000K iterations of online adaptation~in~\%. [S],~[A] = average error for 5 and 7 categories.}
\vspace{-1mm}
\scalebox{0.77}[0.77]{
\label{tab:online_eval}
\begin{tabular}{l|c|c|c|c|c|c|c}
Method  &  5s     &  10s  &  20s     &  40s     &  80s    &  120s   &  160s \\[0mm] \\
\hline\hline
[S] \textbf{OCN (0-50)}        &  18.9   &  18.0  &  15.5  &  13.2  &  9.8  &  6.8  &  6.1 \\[0mm]
[S] \textbf{OCN (0-1000)}      &  17.3   &  14.9  &  13.6  &  11.5  &  8.3  &  3.4  &  3.3  \\[0mm]
[S] ResNet50                   &  54.7   &  54.7  &  54.7  &  54.7 &  54.7 &  54.7 &  54.7 \\[0mm]
\hline
[A] \textbf{OCN (0-50)}        &  29.4  &  28.2 &  26.4  &  25.0  & 23.1  & 21.0  &  19.8 \\[0mm]
[A] \textbf{OCN (0-1000)}      &  27.7  &  25.7 &  24.4  &  22.0  & 19.9  & 15.4  &  15.0 \\[0mm]
[A] ResNet50                   &  63.9  &  63.9 &  63.9  &  63.9  & 63.9  &  63.9  &  63.9 \\[0mm]
\end{tabular}
}
\end{center}
\vspace{-3mm}

\end{table}

Fig.~\ref{fig:view_to_view} illustrates how objects of one view (anchors) are matched to the objects of another view. We can find the nearest neighbors (positives) in the scene through the OCN embedding space as well as the closest matching objects with descending similarity (negatives). For our synthetic data we report the quality of finding corresponding objects in Tab.~\ref{tab:view_to_view} and differentiate between `attribute errors', that indicate a mismatch of specific attributes (e.g. a blue cup is associated to a red cup), and `object matching errors', which measure when objects are not of the same instance. An OCN embedding significantly improves detecting object instances across multiple views.

\begin{table}[h]
\begin{center}
\caption{Object correspondences errors: attribute error indicates a mismatch of an object attribute, while an object matching error is measured when the matched objects are not the same instance. }
\vspace{-2mm}
\scalebox{0.84}{
\label{tab:view_to_view}
\begin{tabular}{l|c|c}
Method                   & Attribute Error  & Object Matching Error  \\
\hline\hline
OCN supervised           & 4.53\%           & 16.28\%           \\
OCN unsupervised         & 5.27\%           & 18.15\%           \\
Resnet50 embeddings      & 19.27\%          & 57.04\%           \\
\end{tabular}
}
\end{center}
\vspace{-5mm}
\end{table}

\vspace{-1mm}
\subsection{Robot Experiment}
\label{sec:robot_experiments}

\begin{figure}[t]
  \begin{center}
  \includegraphics[width=\linewidth]{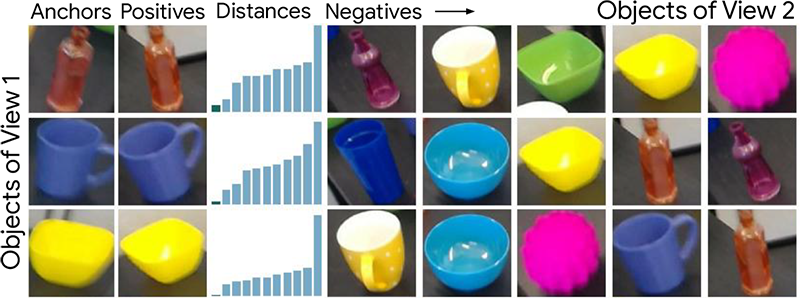}
  \end{center}
  \vspace{-5mm}
  \caption{View-to-view object correspondences: the first column shows all objects detected in one frame (anchors). Each object is associated to the objects found in the other view, objects in the second column are the nearest neighbors. The third column shows the embedding space distance of objects. The other objects are shown from left to right in descending order according to their distances to the anchor (not all objects shown). }
  \label{fig:view_to_view}
\vspace{-2mm}
  \begin{center}
  \includegraphics[width=\linewidth]{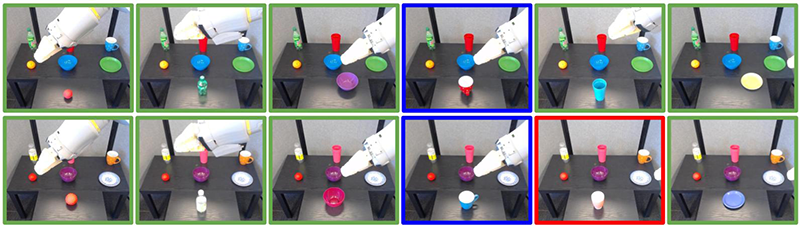}
  \end{center}
  \vspace{-5mm}
  \caption{The robot experiment of pointing to the best match of a query object (placed in front of the robot on the small table). The closest match is selected from two sets of target objects, placed on the table behind the query object. The first and the second row correspond to the experiment for the first and second target set. Images with green frame indicate cases where both the `class' and `container' attributes are matched correctly. Blue frames show where only the `container' attribute is matched correctly and red frames indicate neither attribute is matched.}
  \label{fig:pointing_experiment_short}
\vspace{-4mm}
\end{figure}

\begin{figure*}[ht]
  \begin{center}
  \includegraphics[width=0.6\linewidth]{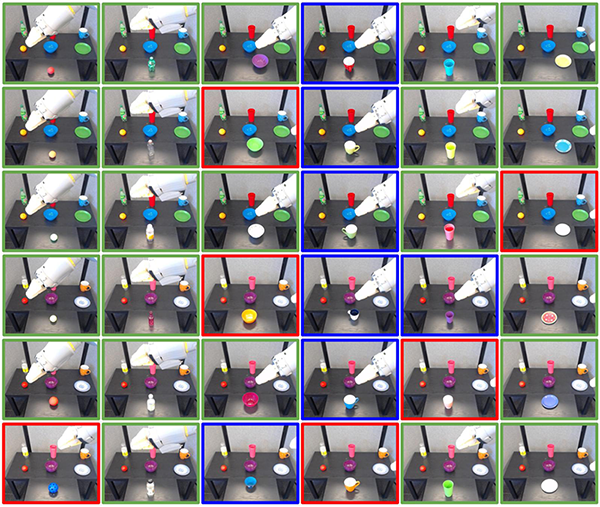}
  \end{center}
  \vspace{-4mm}
  \caption{The robot experiment of pointing to the best match to a query object (placed in front of the robot on the small table). The closest match is selected from two sets of target objects, which are placed on the rear table behind the query object. The first and the last three rows correspond to the experiment for the first and second target object set. Each column also illustrates the query objects for each object category. Images with green frame correspond to cases where both the `class' and `container' attributes are matched correctly. Images with blue frame refer to the cases where only `container' attribute is matched correctly. Images with red frames indicates neither of attributes are matched.}
  \label{fig:pointing_experiment}
  \begin{center}
  \includegraphics[width=0.8\linewidth]{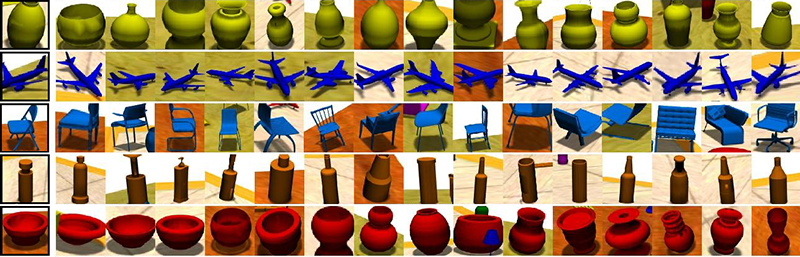}
  \includegraphics[width=0.8\linewidth]{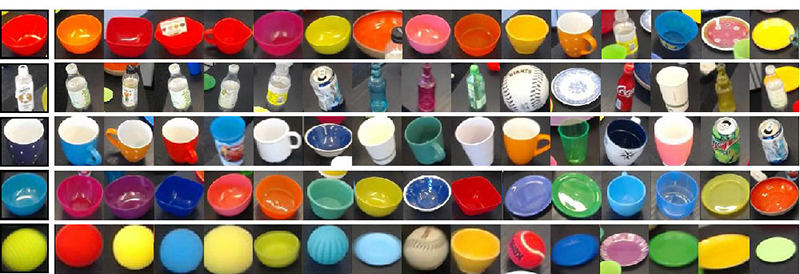}
  \end{center}
  \vspace{-4mm}
  \caption{An OCN embedding organizes objects along their visual and semantic features. For example, a red bowl as query object is associated with other similarly colored objects and other containers. The leftmost object (black border) is the query object and its nearest neighbors are listed in descending order. The top row shows renderings of our synthetic dataset, while the bottom row shows real objects. For real objects we removed the same instance manually.}
  \label{fig:object_query6}
\end{figure*}

To evaluate OCN for real world robotics scenarios we defined a robotics pointing task. The goal of the task is to enable a robot to point to an object that it deems most similar to the object directly in front of it (\fig{pointing_experiment_short}). The objects on the rear table are randomly selected from the object categories (\tabl{real_data_set}). We consider two sets of these target objects. The quantitative experiment in \tabl{pointing_evaluation} uses three query objects per category and is ran three times for each combination of query and target objects ($3 \times 2 \times 18 = 108$ experiments performed). The full set for one of the three runs is shown in \fig{pointing_experiment}.

A quantitative evaluation of OCN performance for this experiment is shown in \tabl{pointing_evaluation}. We report on errors related to `class' and `container' attributes. While the trained OCN model is performing well on the most categories, it has  difficulty on the object classes `cups \& mugs' and `glasses'. These categories are generally mistaken with the category `bowls'. As a result the network performs much better in the attribute `container' since all the three categories `bowls', `bottles \& cans', and 'glasses' refer to the same attribute.

At the beginning of each experiment the robot captures a snapshot of the scene. We then split the captured image into two images: the upper portion of the image that contains the target object set and the lower portion of the image that only contains the query object.
We detect the objects and find the nearest neighbor of the query object in the embedding space to find the closest match.

\subsection{Object Attribute Classification}

One way to evaluate the quality of unsupervised embeddings is to train attribute classifiers on top of the embedding using labeled data. Note however, that this may not entirely reflect the quality of an embedding because it is only measuring a discrete and small number of attributes while an embedding may capture more continuous and larger number of abstract concepts.

\textbf{Classifiers:} we consider two types of classifiers to be applied on top of existing embeddings in this experiment: linear and nearest-neighbor classifiers. The linear classifier consists of a single linear layer going from embedding space to the 1-hot encoding of the target label for each attribute. It is trained with a range of learning rates and the best model is retained for each attribute. The nearest-neighbor classifier consists of embedding an entire `training' set, and for each embedding of the evaluation set, assigning to it the labels of the nearest sample from the training set. Nearest-neighbor classification is not a perfect approach because it does not necessarily measure generalization as linear classification does and results may vary significantly depending on how many nearest neighbors are available. It is also less subject to data imbalances. We still report this metric to get a sense of its performance because in an unsupervised inference context, the models might be used in a nearest-neighbor fashion (e.g. as in \sect{robot_experiments}).

\begin{figure}[t]
  \begin{center}
  \includegraphics[width=\linewidth]{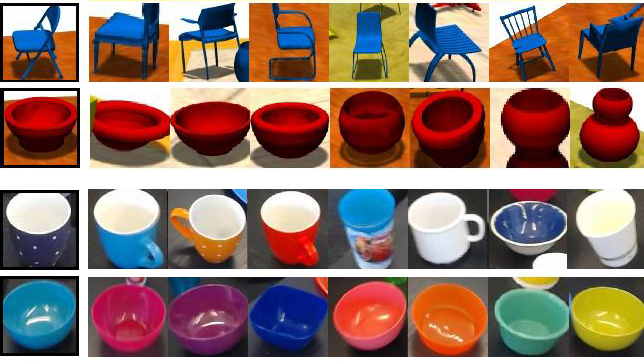}
  \end{center}
  \caption{An OCN embedding organizes objects along their visual and semantic features. For example, a red bowl as query object is associated with other similarly colored objects and other containers. The leftmost object (black border) is the query object and its nearest neighbors are listed in descending order. The top row shows renderings of our synthetic dataset, while the bottom row shows real objects. Please note that these are the nearest neighbors among all objects in the respective dataset.}
   \label{fig:object_query0}
\end{figure}

\textbf{Baselines:} we compare multiple baselines (BL) in \tabl{classification} and \tabl{random_weights}. The `Softmax' baseline refers to the model described in \fig{models}, i.e. the exact same architecture as for OCN except that the model is trained with a supervised cross-entropy/softmax loss. The `ResNet50' baseline refers to using the unmodified outputs of the ResNet50 model~\cite{He2016DeepRL} (2048-dimensional vectors) as embeddings and training a nearest-neighbor classifier as defined above. We consider `Softmax' and `ResNet50' baselines as the lower and upper error-bounds for standard approaches to a classification task. The `OCN supervised' baseline refers to the exact same OCN training described in \fig{models}, except that the positive matches are provided rather than discovered automatically. `OCN supervised' represents the metric learning upper bound for classification. Finally we indicate as a reference the error rates for random classification.

\begin{table}[t]
\begin{center}
\caption{\textbf{Attributes classification errors:} using attribute labels, we train either a linear or nearest-neighbor classifier on top of existing fixed embeddings. The supervised OCN is trained using labeled positive matches, while the unsupervised one decides on positive matches on its own. All models here are initialized and frozen with ImageNet-pretrained weights for the ResNet50 part of the architecture, while the additional layers above are random and trainable.}
\vspace{-5mm}
\scalebox{0.68}{
\label{table:classification}
\begin{tabular}{l|c|c|c|c}
                                                        & Class (12)  & Color (8)     & Binary        &           \\
                                                        & Attribute   & Attribute     & Attributes    & Embedding \\
Method                                                  & Error       & Error         & Error         & Size      \\
\hline
\hline
[BL] Softmax                                      & 2.98\%      & 0.80\%        & 7.18\%        & -         \\[0mm]
[BL] OCN sup (linear)                      & 7.49\%      & 3.01\%        & 12.77\%       & 32        \\[0mm]
[BL] OCN sup (NN)                          & 9.59\%      & 3.66\%        & 12.75\%       & 32        \\[0mm]
\textbf{[ours] OCN unsup. (linear)}    & 10.70\%     & 5.84\%        & 13.76\%       & 24        \\[0mm]
\textbf{[ours] OCN unsup. (NN)}        & 12.35\%     & 8.21\%        & 13.75\%       & 24        \\[0mm]
[BL] ResNet50 embed. (NN)                     & 14.82\%     & 64.01\%       & 13.33\%       & 2048      \\[0mm]
[BL] Random Chance                                       & 91.68\%     & 87.50\%       & 50.00\%       & -         \\[0mm]
\end{tabular}
}
\end{center}
\vspace{-6mm}
\end{table}

\textbf{Results:} we quantitatively evaluate our unsupervised models against supervised baselines on the labeled synthetic datasets (train and test) introduced in \sect{synthetic_data}. Note that there is no overlap in object instances between the training and the evaluation set.
The first take-away is that unsupervised performance closely follows its supervised baseline when trained with metric learning.
As expected the cross-entropy/softmax approach performs best and establishes the error lower bound while the ResNet50 baseline are upper-bound results.
In \fig{object_query0} and Sec.~\ref{sec:appendix_qualitative}, we show  results of nearest neighbor objects discovered by OCN.

\section{Conclusion and Future Work}

We introduced a self-supervised objective for object representations able to disentangle object attributes, such as color, shape, and function.
We showed this objective can be used in online settings which is particularly useful for robotics to increase robustness and adaptability to unseen objects. We demonstrated a robot is able to discover similarities between objects and pick an object that most resembles one presented to it.
In summary, we find that within a single scene with novel objects, the more our model looks at objects, the more it can recognize them and understand their visual attributes, despite never receiving any labels for them.
\\  
Current limitations include relying on all objects to be present in all frames of a video. Relaxing this limitation will allow to use the model in unconstrained settings.
Additionally, the online training is currently not real-time as we first set out to demonstrate the usefulness of online-learning in non-real-time.
Real-time training requires additional engineering that is beyond the scope of this research.
Finally, the model currently relies on an off-the-self object detector which might be noisy, an avenue for future research is to back-propagate gradients through the objectness model to improve detection and reduce noise.

{\small
\bibliographystyle{ieee}
\bibliography{egbib}
}

\clearpage
\newpage
\section*{Supplementary Material}

In the following we provide details on our datasets and report additional experiments and results.

\section{Dataset}

\begin{table}[ht]
\begin{center}
\caption{Details on our three datasets: head-mounted videos for online training, automatically captured by a robot, and synthetic. }
\vspace{-2mm}
\scalebox{0.85}{
\label{tab:data_set}
\begin{tabular}{l|c|c|c|c}
                &              & \#Unique           &          & \#Views/Frames              \\
Dataset         & \#Categories & \#Objects          & $\#$Scenes & per Scene \\
\hline\hline
$Real_{head}$     & 7            & 75$+$      & 25       & 3000             \\
$Real_{auto}$     & 6            & 187      & 576      & 115$-$230          \\
$Synthetic$      & 12           & 4k       & 250k     & 2                 \\
\end{tabular}
}
\end{center}
\vspace{-2mm}
\begin{center}
\caption{Real object dataset: we use 187 unique object instances spread across six categories.}
\scalebox{0.68}{
\label{table:real_data_set}
\begin{tabular}{l|c|c|c|c|c|c}
            & Balls & Bottles \& Cans & Bowls & Cups \& Mugs & Glasses & Plates \\ \hline \hline
Training    & 14    & 13              & 19    & 19           & 22      & 23     \\
Validation  & 5     & 4               & 8     & 6            & 5       & 6      \\
Test        & 6     & 6               & 10    & 6            & 6       & 9      \\ \hline
Total       & 25    & 23              & 37    & 31           & 33      & 38     \\
\end{tabular}
}
\end{center}
\vspace{-4mm}
\end{table}

\section{Random Weights}
\label{sec:random_weights}

We find in \tabl{random_weights} that models that are not pretrained with ImageNet supervision perform worse but still yield reasonable results. This indicates that the approach does not rely on a good initialization to bootstrap itself without labels. Even more surprisingly, when freezing the weights of the ResNet50 base of the model to its random initialization, results degrade but still remain far below chance as well as below the 'ResNet50 embeddings' baseline. Obtaining reasonable results with random weights has already been observed in prior work such as \cite{jarrett2009best}, \cite{saxe2011random} and \cite{DBLP:journals/corr/abs-1711-10925}.

\begin{table}[ht]
\begin{center}
\caption{\textbf{Results with random weights} (no ImageNet pre-training)}
\vspace{-4mm}
\scalebox{0.68}{
\label{table:random_weights}
\begin{tabular}{l|c|c|c|c}
                                                     & Class (12)     & Color (8)     & Binary        &                  \\
                                                     & Attribute      & Attribute     & Attributes    &                   \\
Method                                               & Error          & Error         & Error         & Finetuning        \\
\hline
\hline
[BL] Softmax                                    & 23.18\%       & 10.72\%       & 13.56\%            & yes             \\[0mm]
[BL] OCN sup. (NN)                        & 29.99\%       & 2.23\%        & 20.25\%            & yes             \\[0mm]
[BL] OCN sup. (linear)                    & 34.17\%       & 2.63\%        & 27.37\%            & yes             \\[0mm]
\textbf{[ours] OCN unsup. (NN)}      & 35.51\%       & 2.93\%        & 22.59\%            & yes              \\[0mm]
\textbf{[ours] OCN unsup. (linear)}  & 47.64\%       & 4.43\%        & 35.73\%            & yes              \\[0mm]
\hline
[BL] Softmax                                    & 27.28\%       & 5.48\%        & 20.40\%            & no             \\[0mm]
[BL] OCN sup. (NN)                        & 37.90\%       & 4.00\%        & 23.97\%            & no             \\[0mm]
[BL] OCN sup. (linear)                    & 39.98\%       & 4.68\%        & 32.74\%            & no             \\[0mm]
\textbf{[ours] OCN unsup. (NN)}      & 43.01\%       & 5.56\%        & 26.29\%            & no              \\[0mm]
\textbf{[ours] OCN unsup. (linear)}  & 48.26\%       & 6.15\%        & 37.05\%            & no              \\[0mm]
[BL] ResNet50 embed. (NN)                   & 59.65\%       & 21.14\%        & 34.94\%            & no             \\[0mm]
\hline
[BL] Random Chance                          & 91.68\%     & 87.50\%       & 50.00\%       & -         \\[0mm]
\end{tabular}
}
\end{center}
\end{table}

\section{Additional Experiments and Results}
\label{sec:appendix_qualitative}

\begin{table}[ht]

\begin{center}
\caption{
    Evaluation of robotic pointing: we report on two attribute errors: `class' and `container'. An error for 'class' is reported when the robot points to an object of a different class among the 5 categories: balls, plates, bottles, cups, bowls. An error for 'container' is reported when the robot points to a non-container object when presented with a container object, and vice-versa.
    }

\vspace{2mm}
\label{table:pointing_evaluation}
\begin{tabular}{l|l|l}
Objects        & Class Error            & Container Error \\
\hline\hline
Balls             & 11.1 $\pm$  7.9\%         & 11.1 $\pm$ 7.9\%\\
Bottles \& Cans   & 0.0 $\pm$  0.0\%          & 0.0 $\pm$ 0.0\% \\
Bowls             & 22.2 $\pm$ 15.7\%        & 16.7 $\pm$ 0.0\% \\
Cups \& Mugs      & 88.9 $\pm$ 7.9\%         & 16.7 $\pm$ 13.6\% \\
Glasses           & 38.9 $\pm$ 7.9\%         & 5.6 $\pm$ 7.9\% \\
Plates            & 5.6 $\pm$ 7.9\%          & 11.1 $\pm$ 2.3\% \\
\hline
Total             & 27.8 $\pm$ 3.9\%         & 11.1 $\pm$ 2.3\% \\

\end{tabular}
\end{center}
\vspace{-6mm}
\end{table}

\begin{figure*}
  \begin{center}
  \includegraphics[width=\linewidth]{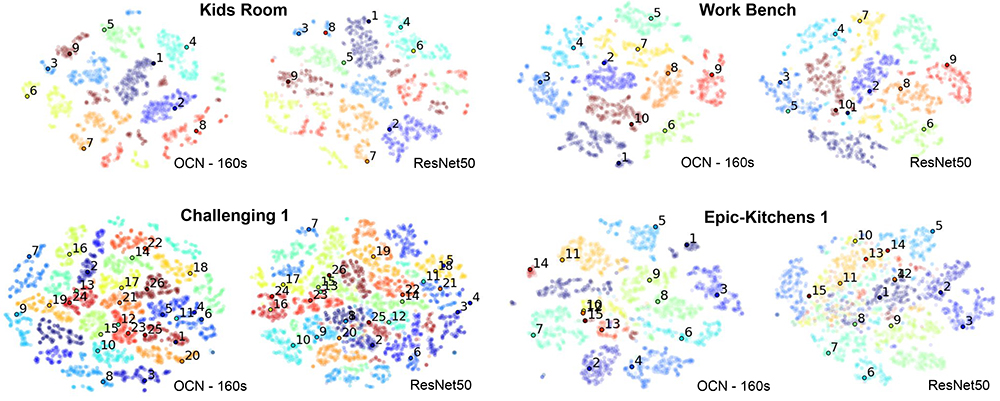}
  \end{center}
  \vspace{-4mm}
  \caption{T-SNE plots of different video sequences. The plots show each object of the 600 frames used for evaluation with their ground truth index as color. Compared to the ResNet50 baseline, OCN trained on 160 seconds of video produces more pronounced clusters which indicates an improved disentanglement of object features. Reference objects are shown with a black border and an index in the order of their appearance in Figure~\ref{fig:all_videos}. We used 64-dimensional OCN embeddings and 2048-dimensional ResNet50 embeddings. All plots where generated with a perplexity value of 12. }
  \label{fig:tsne}
\end{figure*}

\begin{figure*}[ht]
  \begin{center}
  \includegraphics[width=\linewidth]{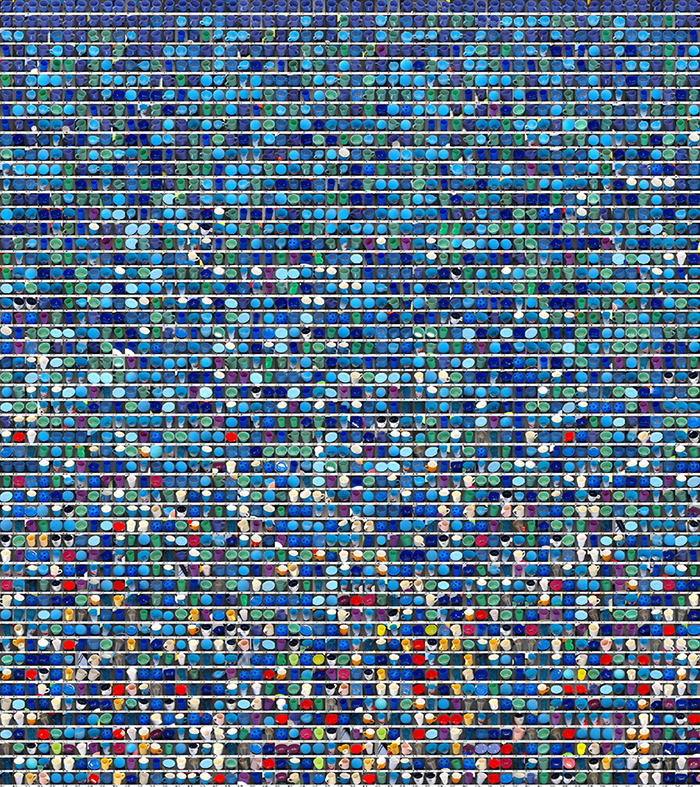}
  \end{center}
  \caption{A result showing the organization of real bowls based on OCN embeddings. The query object (black border, top left) was taken from the validation all others from the training data. As the same object is used in multiple scenes the same object is shown multiple times.}
  \label{fig:object_query2}
\end{figure*}
\begin{figure*}[ht]
  \begin{center}
  \includegraphics[width=\linewidth]{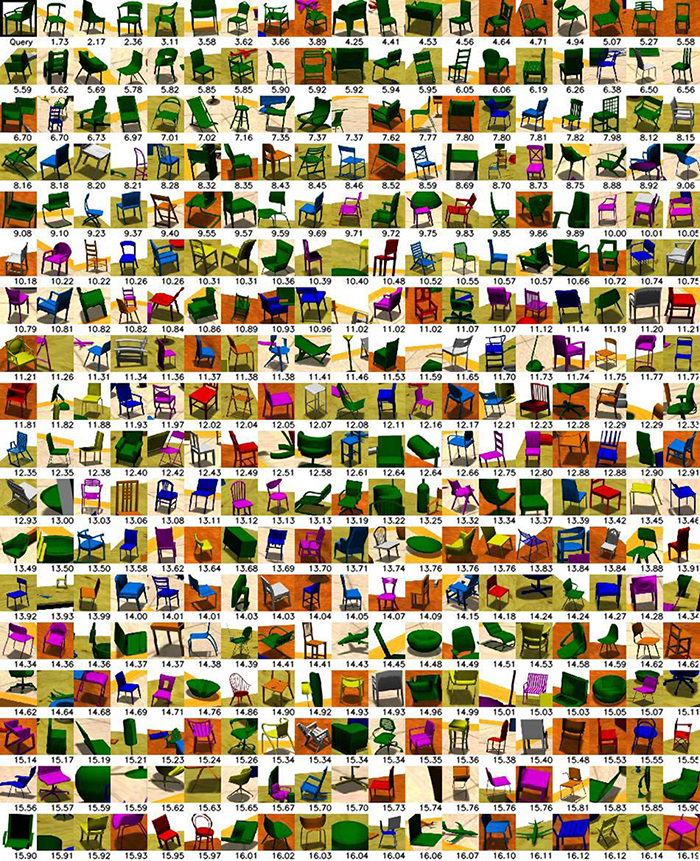}
  \end{center}
  \caption{A result showing the organization of chairs from synthetic data based on OCN embeddings. The query object (black border, top left) was taken from the validation all others from the training data.}
  \label{fig:object_query5}
\end{figure*}

\end{document}